\documentclass[10pt,twocolumn,letterpaper]{article}

\usepackage{cvpr}
\usepackage{times}
\usepackage{epsfig}
\usepackage{graphicx}
\usepackage{amsmath}
\usepackage{amssymb}
\usepackage{caption}
\usepackage{multirow}
\usepackage{tabu}
\usepackage[symbol]{footmisc}
\usepackage[breaklinks=true,bookmarks=false,hyperfootnotes=false]{hyperref}

\usepackage{etoolbox}
\patchcmd{\footref}{\ref}{\ref*}{}{}

\newcommand\T{\rule{0pt}{2.9ex}}       % Top strut
\newcommand\B{\rule[-1.2ex]{0pt}{0pt}} % Bottom strut

\cvprfinalcopy % *** Uncomment this line for the final submission
\renewcommand{\thefootnote}{\fnsymbol{footnote}}

\newcommand\blfootnote[1]{%
	\begingroup
	\renewcommand\thefootnote{}\footnote{#1}%
	\addtocounter{footnote}{-1}%
	\endgroup
}

\def\cvprPaperID{1638} % *** Enter the CVPR Paper ID here

\begin{document}

%%%%%%%%% TITLE
\title{\textit{Veritatem Dies Aperit} - Temporally Consistent Depth Prediction Enabled by a Multi-Task Geometric and Semantic Scene Understanding Approach}

\author{Amir Atapour-Abarghouei$^1$ \quad Toby P. Breckon$^{1,2}$\\
	$^{1}$Department of Computer Science -- $^{2}$Department of Engineering\\
	Durham University, UK\\
	{\tt\small \{amir.atapour-abarghouei,toby.breckon\}@durham.ac.uk}}

\maketitle
\thispagestyle{empty}

%%%%%%%%% ABSTRACT
\begin{abstract}
   Robust geometric and semantic scene understanding is ever more important in many real-world applications such as autonomous driving and robotic navigation. In this paper, we propose a multi-task learning-based approach capable of jointly performing geometric and semantic scene understanding, namely depth prediction (monocular depth estimation and depth completion) and semantic scene segmentation. Within a single temporally constrained recurrent network, our approach uniquely takes advantage of a complex series of skip connections, adversarial training and the temporal constraint of sequential frame recurrence to produce consistent depth and semantic class labels simultaneously. Extensive experimental evaluation demonstrates the efficacy of our approach compared to other contemporary state-of-the-art techniques.\vspace{-0.50cm}%%
\end{abstract}\vspace{-0.15cm}

\section{Introduction}
\label{sec:intro}\vspace{-0.1cm}

As scene understanding\blfootnote{\textit{Veritatem Dies Aperit}: Time discovers the truth.} grows in popularity due to its applicability in many areas of interest for industry and academia, scene depth has become ever more important as an integral part of this task. Whilst in many current autonomous driving solutions, imperfect stereo camera set-ups or expensive LiDAR sensors are used to capture depth, research has recently focused on refining estimated depth with corrupted or missing regions in post-processing, rendering it more useful in any downstream applications \cite{abarghouei18review, xue2017depth, zhang2018deep}. Moreover, monocular depth estimation has received significant attention within the research community as a cheap and innovative alternative to other more expensive and performance-limited technologies \cite{atapour2018real, eigen2015predicting, monodepth17, zhou2017unsupervised}.\blfootnote{\href{https://github.com/atapour/temporal-depth-segmentation}{Code}: \href{https://github.com/atapour/temporal-depth-segmentation}{https://github.com/atapour/temporal-depth-segmentation}.}%

Pixel-level image understanding, namely semantic segmentation, also plays an important role in many vision-based systems. Significant success has been achieved using Convolutional Neural Networks (CNN) in this field \cite{badrinarayanan2015segnet, chen2018deeplab, long2015fully, ronneberger2015u, stollenga2015parallel} and many others such as image classification \cite{maggiori2017convolutional}, object detection \cite{zhu2017flow} and alike in recent years.%
%........................................................
\begin{figure}[t!]
	\centering
	\includegraphics[width=0.99\linewidth]{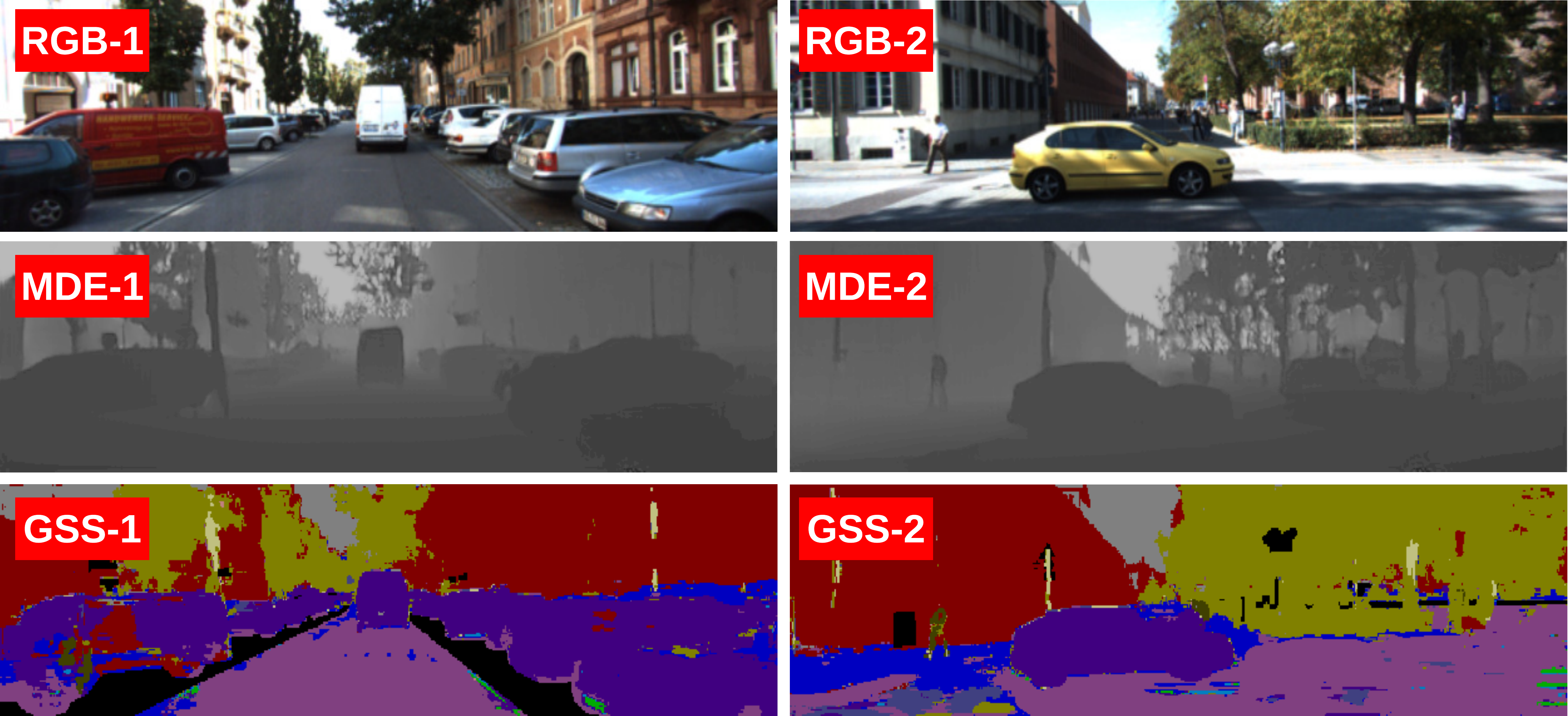}
	\captionsetup[figure]{skip=7pt}
	\captionof{figure}{Exemplar results of the proposed approach. \textbf{RGB:} input colour image; \textbf{MDE:} Monocular Depth Estimation; \textbf{GSS:} Generated Semantic Segmentation.}%
	\label{fig:taste}\vspace{-0.7cm}
\end{figure}
%........................................................

\begin{figure*}[t!]
	\centering
	\includegraphics[width=0.99\linewidth]{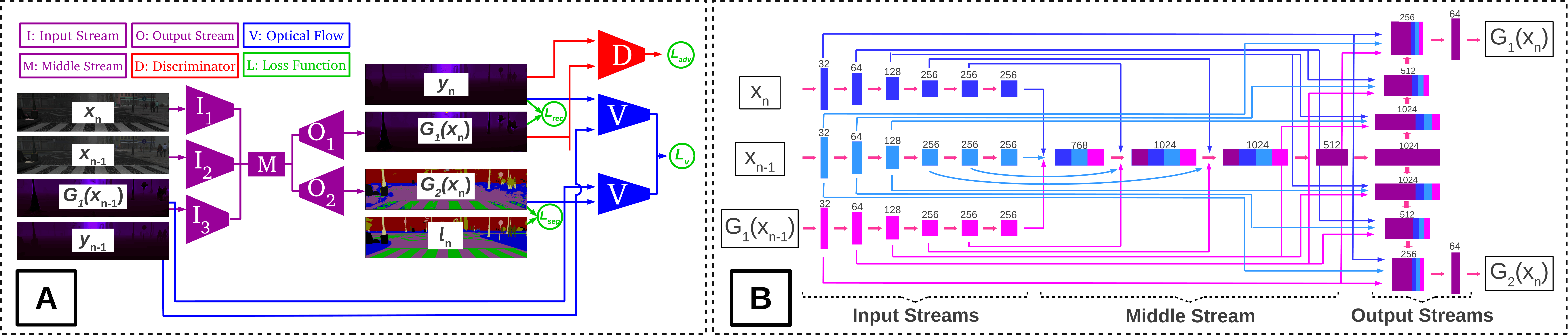}
	\captionsetup[figure]{skip=7pt}
	\captionof{figure}{Overall training procedure of the model (A) and the detailed outline of the generator architecture (B).}%
	\label{fig:pipeline-architecture}\vspace{-0.5cm}
\end{figure*}

In this work, we propose a model capable of semantically understanding a scene by jointly predicting depth and pixel-wise semantic classes (Figure \ref{fig:taste}). The network performs semantic segmentation (Section \ref{ssec:segmentation_approach}) along with monocular depth estimation (\ie, predicting scene depth based on a single RGB image) or depth completion (\ie, completing missing regions of existing depth sensed through other imperfect means, Section \ref{ssec:depth_approach}). Our approach performs these tasks within a single model (Figure \ref{fig:pipeline-architecture} (A)) capable of two separate scene understanding objectives requiring low-level feature extraction and high-level inference, which leads to improved and deeper representation learning within the model \cite{kendall2018multi}. This is empirically demonstrated via the notably improved results obtained for each individual task when performed simultaneously in this manner.%
%........................................................
%........................................................

Within the current literature, many techniques focus on individual frames to spatially accomplish their objectives, ignoring temporal consistency in video sequences, one of the most valuable sources of information widely available within real-world applications. In this work, we propose a feedback network that at each time step takes the output generated at the previous time step as a recurrent input. Furthermore, using a pre-trained optical flow estimation model, we ensure the temporal information is explicitly considered by the overall model during training (Figure \ref{fig:pipeline-architecture} (A)).%

In recent years, skip connections have been proven to be very effective when the input and output of a CNN share similar high-level spatial features \cite{orhan2017skip, ronneberger2015u, tong2017image, yamanaka2017fast}. We make use of a complex network of skip connections throughout the architecture to guarantee that no high-level spatial features are lost during training as the features are down-sampled. In short, our main contributions are as follows:\vspace{-0.12cm}
\begin{itemize} 
	\setlength\itemsep{1.2mm}
	\item \textit{Depth Prediction} - via a supervised multi-task model adversarially trained using complex skip connections that can predict depth (monocular depth estimation and depth completion) having been trained on high-quality synthetic training data \cite{RosCVPR16} (Section \ref{ssec:depth_approach}).\vspace{-0.22cm}
	\item \textit{Semantic Segmentation} - via the same multi-task model, which is capable of performing the task of semantic scene segmentation as well as the aforementioned depth estimation/completion (Section \ref{ssec:segmentation_approach}).\vspace{-0.22cm}
	\item \textit{Temporal Continuity} - temporal information is explicitly taken into account during training using both recurrent network feedback and gradients from a pre-trained frozen optical flow network.\vspace{-0.22cm}
\end{itemize}

This leads to a novel scene understanding approach capable of temporally consistent geometric depth prediction and semantic scene segmentation whilst outperforming prior work across the domains of monocular depth estimation \cite{atapour2018real, eigen2014depth, monodepth17, liu2016learning, zhan2018unsupervised, zhou2017unsupervised}, 
completion \cite{abarghouei16filling, iizuka2017globally, liu2012guided, yu2018generative} and semantic segmentation \cite{badrinarayanan2015segnet, chen2018deeplab, kendall2017bayesian, liu2015semantic, long2015fully, noh2015learning, uhrig2016pixel, visin2016reseg, zheng2015conditional}.\vspace{-0.05cm}%

%-------------------------------------------------------------------------
\section{Related Work}
\label{sec:related}\vspace{-0.10cm}

We consider relevant prior work over three distinct areas, semantic segmentation (Section \ref{ssec:semantic_related}), monocular depth estimation (Section \ref{ssec:estimation_related}), and depth completion (Section \ref{ssec:completion_related}).\vspace{-0.05cm}%

\subsection{Semantic Segmentation}
\label{ssec:semantic_related}\vspace{-0.1cm}

Within the literature, promising results have been achieved using fully-convolutional networks \cite{long2015fully}, saved pooling indices \cite{badrinarayanan2015segnet}, skip connections \cite{ronneberger2015u}, multi-path refinement \cite{lin2017refinenet}, spatial pyramid pooling \cite{zhao2017pyramid}, attention modules focusing on scale or channel \cite{chen2016attention, yu2018learning} and others.%
%............................................................................
\begin{table*}[!t]
	\centering
	\resizebox{\textwidth}{!}{
		{\tabulinesep=0mm
			\begin{tabu}{@{\extracolsep{5pt}}c c c c c c c c c c}
				\hline\hline
				\multicolumn{1}{l}{\multirow{2}{*}{Method}} &  
				\multicolumn{4}{c}{Depth Error (lower, better)} & 
				\multicolumn{3}{c}{Depth Accuracy (higher, better)} &
				\multicolumn{2}{c}{Segmentation (higher, better)} \T\B \\
				\cline{2-5} \cline{6-8} \cline{9-10}
				& Abs. Rel. & Sq. Rel. 	& RMSE 	& RMSE log 	& $\sigma < 1.25$& $\sigma < 1.25^{2}$& $\sigma < 1.25^{3}$& Accuracy& IoU \T\B \\
				\hline
				Two Models		& 0.245 	& 1.513 	& 6.323 & 0.274 	& 0.803 		 & 0.856 			  & 0.882 			   & 0.604 	 & 0.672 \T \\
				One Model 	& 0.208 	& 1.402 	& 6.026 & 0.269 	& 0.836 		 & 0.901 			  & 0.926 			   & 0.748 	 & 0.764 \B \\
				\hline
			\end{tabu}}}
	\captionsetup[table]{skip=7pt}
	\captionof{table}{Comparison of depth prediction and segmentation tasks performed in one single network and two separate networks.}
	\label{table:ablation_separation}\vspace{-0.25cm}
\end{table*}
%............................................................................
%........................................................
\begin{figure*}[!t]
	\centering
	\includegraphics[width=0.99\linewidth]{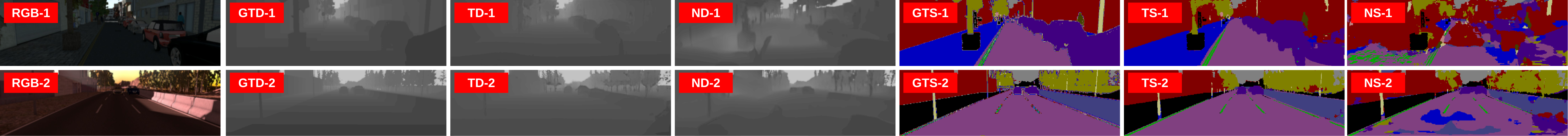}
	\captionsetup[figure]{skip=7pt}
	\captionof{figure}{Comparing the results of the approach on synthetic test set when the model is trained with and without temporal consistency. \textbf{RGB:} input colour image; \textbf{GTD:} Ground Truth Depth; \textbf{GTS:} Ground Truth Segmentation; \textbf{TS:} Temporal Segmentation; \textbf{TD:} Temporal Depth; \textbf{NS:} Non-Temporal Segmentation; \textbf{ND:} Non-Temporal Depth.}%
	\label{fig:temporal}\vspace{-0.6cm}
\end{figure*}
%........................................................

Temporal information in videos has also been used to improve segmentation accuracy or efficiency. \cite{fayyaz2016stfcn} proposes a spatio-temporal LSTM based on frame features for higher accuracy. Labels are propagated in \cite{nilsson2016semantic} using gated recurrent units. In \cite{gadde2017semantic}, features from preceding frames are warped via flow vectors to reinforce the current frame features. On the other hand, \cite{shelhamer2016clockwork} reuses previous frame features to reduce computation. In \cite{zhu17dff}, an optical flow network \cite{dosovitskiy2015flownet} is used to propagate features from key frames to the current one. Similarly, \cite{xu2018dynamic} uses an adaptive key frame scheduling policy to improve both accuracy and efficiency. Additionally, \cite{li2018low} proposes an adaptive feature propagation module that employs spatially variant convolutions to fuse the frame features, thus further improving efficiency. Even though the main objective of this work is not semantic segmentation, it can be demonstrated that when the main objective (depth prediction) is performed alongside semantic segmentation, the results are superior to when the tasks are performed individually (Table \ref{table:ablation_separation}).\vspace{-0.15cm}%

\subsection{Monocular Depth Estimation}
\label{ssec:estimation_related}\vspace{-0.1cm}

Estimating depth from a single colour image is very desirable as unlike stereo correspondence \cite{scharstein2002taxonomy}, structure from motion \cite{cavestany15robot} and alike \cite{abrams2012heliometric, tao2015depth}, it leads to a system with reduced size, weight, power and computational requirements. For instance, \cite{baig2014im2depth} employs sparse coding to estimate depth, while \cite{eigen2015predicting, eigen2014depth} generates depth from a two-scale network trained on RGB and depth. Other supervised models such as \cite{laina2016deeper, li2015depth} have also achieved impressive results despite the scarcity of ground truth depth for supervision.%

Recent work has led to the emergence of new techniques that calculate disparity by reconstructing corresponding views within a stereo correspondence framework without ground truth depth. The work by \cite{xie2016deep3d} learns to generate the right view from the left image used as the input while producing an intermediary disparity map. Likewise, \cite{monodepth17} uses bilinear sampling \cite{jaderberg2015spatial} and left/right consistency incorporated into training for better results. In \cite{zhou2017unsupervised}, depth and camera motion are estimated by training depth and pose prediction networks, indirectly supervised via view synthesis. The model in \cite{kuznietsov2017semi} is supervised by sparse ground truth depth and the model is then enforced within a stereo framework via an image alignment loss to output dense depth.%

Additionally, contemporary supervised approaches such as \cite{atapour2018real} have taken to using synthetic depth data to produce sharp and crisp depth outputs. In this work, we also utilize synthetic data \cite{RosCVPR16} in a directly supervised training framework to perform the task of monocular depth estimation.\vspace{-0.15cm}

\subsection{Depth Completion}
\label{ssec:completion_related}\vspace{-0.1cm}

While colour image inpainting has been a long-standing and well-established field of study \cite{bist, breckon12completion, ding2018perceptually, pathak2016context, bistoseh, yeh2017semantic}, its use within the depth modality is considerably less effective \cite{abarghouei18review}. There have been a variety of depth completion techniques in the literature including those utilizing smoothness priors \cite{herrera2013depth}, exemplar-based depth inpainting \cite{atapour2018extended}, low-rank matrix completion \cite{xue2017depth}, object-aware interpolation \cite{abarghouei17depthcomp}, tensor voting \cite{kulkarni2013depth}, Fourier-based depth filling \cite{abarghouei16filling}, background surface extrapolation \cite{matsuo2015depth, muddala2014depth}, learning-based approaches using deep networks \cite{abarghouei19gan, zhang2018deep}, and alike \cite{bertalmio2001navier, chen2014improved, liu2016building}. However, prior work does not include any work focusing on enforcing temporal continuity in a learning-based approach.\vspace{-0.15cm}%

\section{Proposed Approach}
\label{sec:approach}\vspace{-0.1cm}

Our approach is designed to perform two tasks using a single joint model: depth estimation/completion (Section \ref{ssec:depth_approach}) and semantic segmentation (Section \ref{ssec:segmentation_approach}). This has been made possible using a synthetic dataset \cite{RosCVPR16} in which both ground truth depth and pixel-wise segmentation labels are available for video sequences of urban driving scenarios.\vspace{-0.1cm}%

\subsection{Overall Architecture}
\label{ssec:architecture_approach}\vspace{-0.1cm}

Our single network takes three different inputs producing two separate outputs for two tasks - \textit{depth prediction and semantic segmentation}. Moreover, temporal information is explicit in our formulation, as one of the inputs at every time step is an output from the previous time step via recurrence. The network comprises three different components: the input streams (Figure \ref{fig:pipeline-architecture} (B) - left), in which the inputs are encoded, the middle stream (Figure \ref{fig:pipeline-architecture} (B) - middle), which fuses the features and begins the decoding process, and finally the output streams (Figure \ref{fig:pipeline-architecture} (B) - right), in which the results are generated.%
%........................................................
\begin{figure*}[t!]
	\centering
	\includegraphics[width=0.99\linewidth]{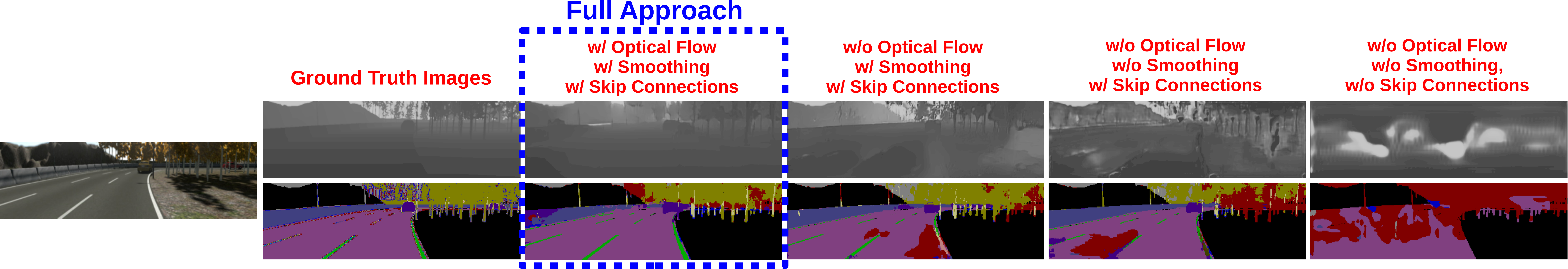}
	\captionsetup[figure]{skip=7pt}
	\captionof{figure}{Comparing the performance of the approach with differing components of the loss function removed.}%
	\label{fig:ablation}\vspace{-0.3cm}
\end{figure*}
%........................................................
%............................................................................
\begin{table*}[!t]
	\centering
	\resizebox{\textwidth}{!}{
		{\tabulinesep=0mm
			\begin{tabu}{@{\extracolsep{5pt}}c c c c c c c c c c}
				\hline\hline
				\multicolumn{1}{l}{\multirow{2}{*}{Method}} &  
				\multicolumn{4}{c}{Depth Error (lower, better)} & 
				\multicolumn{3}{c}{Depth Accuracy (higher, better)} &
				\multicolumn{2}{c}{Segmentation (higher, better)} \T\B \\
				\cline{2-5} \cline{6-8} \cline{9-10}
				& Abs. Rel. 	& Sq. Rel. 		& RMSE 			& RMSE log 		& $\sigma < 1.25$& $\sigma < 1.25^{2}$& $\sigma < 1.25^{3}$& Accuracy& IoU \T\B \\
				\hline\hline
				T/R				& 0.991 		& 1.964 		& 7.393 		& 0.402 		& 0.598 		& 0.684 		& 0.698			& 0.156	& 0.335 \T \\
				T/R/A			& 0.851 		& 1.798 		& 6.826 		& 0.368 		& 0.692 		& 0.750 		& 0.778 		& 0.341	& 0.435	\\
				T/R/A/SC		& 0.655 		& 1.616 		& 6.473 		& 0.278 		& 0.753 		& 0.812 		& 0.838 		& 0.669	& 0.738	\\
				T/R/A/SC/S		& 0.412 		& 1.573 		& 6.256 		& 0.258 		& 0.793 		& 0.875 		& 0.887 		& 0.693	& 0.741	\\
				N/R/A/SC/S		& 0.534			& 1.602 		& 6.469 		& 0.275 		& 0.758			& 0.820			& 0.856 		& 0.614	& 0.681 \B\\
				\hline
				T/R/A/SC/S/OF 	& \textbf{0.208}& \textbf{1.402}& \textbf{6.026}& \textbf{0.269}& \textbf{0.836}& \textbf{0.901}& \textbf{0.926} & \textbf{0.748} & \textbf{0.764} \T\B \\
				\hline
			\end{tabu}}}
	\captionsetup[table]{skip=7pt}
	\captionof{table}{Numerical results with different components of loss. \textbf{T:} Temporal training; \textbf{T:} Non-Temporal training; \textbf{R:} Reconstruction loss; \textbf{A:}  Adversarial loss; \textbf{SC:} Skip Connections; \textbf{S:} Smoothing loss; \textbf{OF:} Optical Flow.}
	\label{table:ablation_remove_components}\vspace{-0.5cm}
\end{table*}
%............................................................................

As seen in Figure \ref{fig:pipeline-architecture} (A), two of the inputs are RGB or RGB-D images (depending on whether monocular depth estimation to create depth, or depth completion to fill holes within an existing depth image, is the focus) from the current and previous time steps. The two input streams that decode these share their weights. The third input is the depth generated at the previous time step. The middle section of the network fuses and decodes the input features and finally the output streams produce the results (scene depth and segmentation). Every layer of the network contains two convolutions, batch normalization \cite{ioffe2015batch} and PReLU \cite{he2015delving}.%

Following recent successes of approaches using skip connections \cite{orhan2017skip, ronneberger2015u, tong2017image, yamanaka2017fast}, we utilize a series of skip connections within our architecture (Figure \ref{fig:pipeline-architecture} (B)). Our inputs and outputs, despite containing different types of information (RGB, depth and pixel-wise class labels), relate to consecutive frames from the same scene and therefore, share high-frequency information such as certain object boundaries, structures, geometry and alike, ensuring skip connections can be of significant value in improving the results. By combining two separate objectives (predicting depth and pixel-wise class labels) within our network, in which the input streams and middle streams are fully trained on both tasks, the results are better than when two separate networks are individually trained to perform the same tasks (Table \ref{table:ablation_separation}).

Even though the entire network is trained as one entity, in our discussions, the parts of the network responsible for predicting depth will be referred to as $G_1$ and the portions involved in semantic segmentation $G_2$. These two modules are essentially the same except for their output streams.\vspace{-0.1cm}% 

\subsection{Depth Estimation / Completion}
\label{ssec:depth_approach}\vspace{-0.1cm}

We consider depth prediction as a supervised image-to-image translation problem, wherein an input RGB image (for depth estimation) or RGB-D image (with the depth channel containing holes for depth completion) is translated to a complete depth image. More formally, a generative model ($G_1$) approximates a mapping function that takes as its input an image $x$ (RGB or RGB-D with holes) and outputs an image $y$ (complete depth image) $G_1 : x \rightarrow y$.%

The initial solution would be to minimize the Euclidean distance between the pixel values of the output ($G_1(x)$) and the ground truth depth ($y$). This simple reconstruction mechanism forces the model to generate images that are structurally and contextually close to the ground truth. For monocular depth estimation, this reconstruction loss is:\vspace{-0.15cm}%
\begin{equation}  
\mathcal{L}_{rec}  = ||G_1(x) - y ||_{1},
\label{lossRec_estimation}\vspace{-0.6cm}
\end{equation}\\
where $x$ is the input image, $G_1(x)$ is the output and $y$ the ground truth. For depth completion, however, the input $x$ is a four-channel RGB-D image with the depth containing holes that would occur during depth sensing. Since we use synthetic data \cite{RosCVPR16}, we only have access to hole-free pixel-perfect ground truth depth. While one could na\"ively cut out random sections of the depth image to simulate holes, as other approaches have done \cite{pathak2016context, yeh2017semantic}, we opt for creating realistic and semantically meaningful holes with characteristics of those found in real-world images \cite{abarghouei18review}. A separate model is thus created and tasked with predicting where holes would be by means of pixel-wise segmentation. A number of stereo images ($30,000$) \cite{Geiger2013IJRR} are used to train the \textit{hole prediction} model by calculating the disparity using Semi-Global Matching \cite{hirschmuller2008stereo} and generating a hole mask ($M$) which indicates which image regions contain holes. The left RGB image is used as the input and the generated mask as the ground truth label, with cross-entropy as the loss function.%

When our main model is being trained to perform depth completion, the hole mask generated by the \textit{hole prediction} network is employed to create the depth channel of the input RGB-D image. Subsequently, the reconstruction loss is:\vspace{-0.10cm}
\begin{equation}  
\mathcal{L}_{rec}  = ||(1 - M) \odot G_1(x) - (1 - M) \odot y ||_{1},
\label{lossRec_compltion}\vspace{-0.5cm}
\end{equation}\\
where $\odot$ is the element-wise product operation and $x$ the input RGB-D image in which the depth channel is $y \odot M$. Experiments with an $L2$ loss returned similar results.%

However, the sole use of a reconstruction loss would lead to blurry outputs since monocular depth estimation and depth completion are multi-modal problems, \ie, several plausible depth outputs can correctly correspond to a region of an RGB image. This multi-modality results in the generative model ($G_1$) averaging all possible modes rather than selecting one, leading to blurring effects in the output. To prevent this, adversarial training \cite{goodfellow2014generative} has become prevalent within the literature \cite{atapour2018real, dosovitskiy2016generating, isola2016image, pathak2016context, yeh2017semantic} since it forces the model to select a mode from the distribution resulting in better quality outputs. In this vein, our depth generation model ($G_1$) takes $x$ as its input and produces fake samples $G_1(x) = \tilde{y}$ while a discriminator ($D$) is adversarially trained to distinguish fake samples $\tilde{y}$ from ground truth samples $y$. The adversarial loss is thus as follows:\vspace{-0.05cm}%
\begin{equation}
\begin{split}  
\mathcal{L}_{adv} = \min_{G_1} \max_{D}\ & \mathop{\mathbb{E}}_{x,y \sim \mathbb{P}_{d}(x,y)} [log D(x,y)] + \\
& \mathop{\mathbb{E}}_{x \sim \mathbb{P}_{d}(x)} [log(1 - D(x, G_1(x)))],
\label{lossAdv}\vspace{-7cm}
\end{split}\vspace{-6cm}
\end{equation}\\
where $\mathbb{P}_{d}$ is the data distribution defined by $\tilde{y} = G_1(x)$, with $x$ being the generator input and $y$ the ground truth.%

Additionally, a smoothing term \cite{monodepth17, heise2013pm} is utilized to encourage the model to generate more locally-smooth depth outputs. Output depth gradients ($\partial G_1(x)$) are penalized using $L1$ regularization, and an edge-aware weighting term based on input image gradients ($\partial x$) is used since image gradients are stronger where depth discontinuities are most likely found. The smoothing loss is therefore as follows:\vspace{-0.05cm}%
\begin{equation}  
\mathcal{L}_{s}  = |\partial G_1(x)| e^{|| \partial x ||},
\label{lossSmooth}\vspace{-0.4cm}
\end{equation}\\
where $x$ is the input and $G_1(x)$ the depth output. The gradients are summed over vertical and horizontal axes.%
%........................................................
\begin{figure}[t!]
	\centering
	\includegraphics[width=0.99\linewidth]{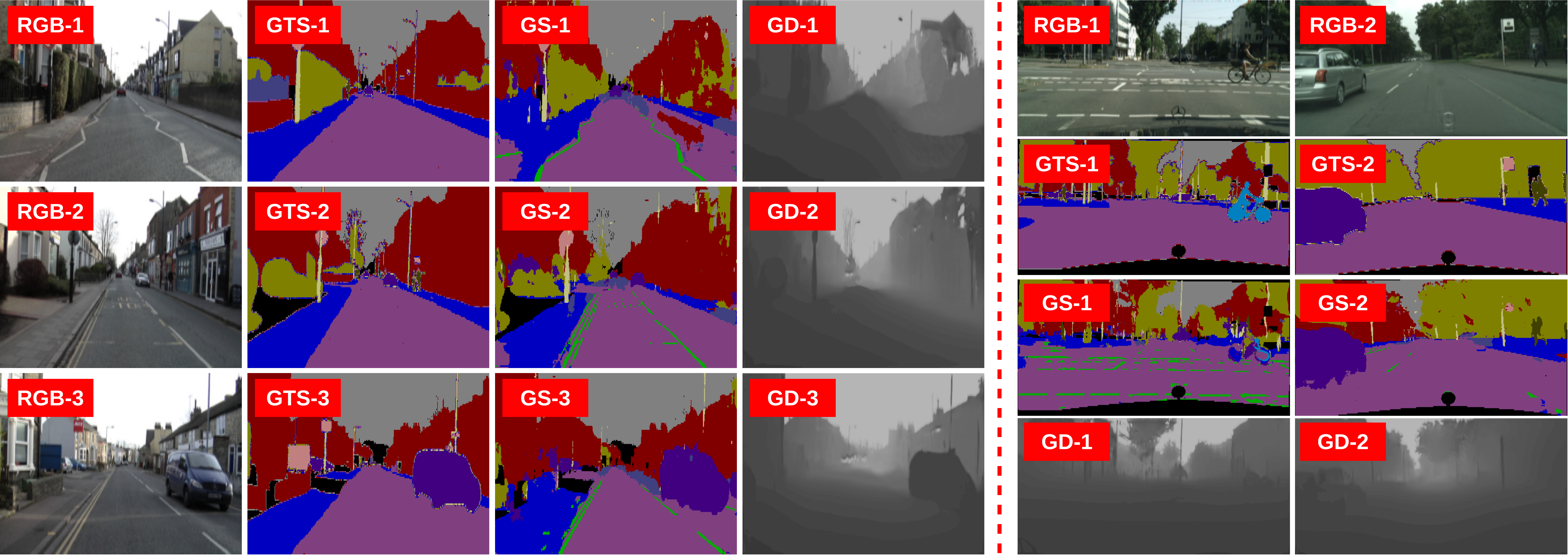}
	\captionsetup[figure]{skip=7pt}
	\captionof{figure}{Results on CamVid \cite{brostow2009semantic} (left) and Cityscapes \cite{cordts2016cityscapes} (right). \textbf{RGB:} input image; \textbf{GTS:} Ground Truth Segmentation; \textbf{GS:} Generated Segmentation; \textbf{GD:} Generated Depth.}%
	\label{fig:segmentation}\vspace{-0.3cm}
\end{figure}
%........................................................
%........................................................
\begin{table}[t!]
	\centering
	\resizebox{0.45\textwidth}{!}{
		\begin{tabular}{c c | c c}
			\hline\hline
			\multicolumn{1}{c}{Method} & \multicolumn{1}{c}{IoU} & \multicolumn{1}{c}{Method} & \multicolumn{1}{c}{IoU} \T\B \\
			\hline
			CRF-RNN \cite{zheng2015conditional} 		& 	62.5	&	DeepLab \cite{chen2018deeplab} 			&	63.1			\T \\	
			Pixel-level Encoding \cite{uhrig2016pixel}	&	64.3	&	FCN-8s \cite{long2015fully} 			&	65.3			\\
			DPN \cite{liu2015semantic}					&	66.8 	&	Our Approach							& \textbf{67.0}		\B \\
			\hline
		\end{tabular}}
	\captionsetup[table]{skip=7pt}
	\captionof{table}{Segmentation on the Cityscapes \cite{cordts2016cityscapes} test set.}
	\label{table:cityscapes}
	\vspace{-0.8cm}
\end{table}
%........................................................ 

Another important consideration is ensuring the depth outputs are temporally consistent. While the model is capable of implicitly learning temporal continuity when the output at each time step is recurrently used as the input at the next time step, we incorporate a light-weight pre-trained optical flow network \cite{ranjan2017optical}, which utilizes a coarse-to-fine spatial pyramid to learn residual flow at each scale, into our pipeline to explicitly enforce consistency in the presence of camera/scene motion. At each time step $n$, the flow between the ground truth depth frames $n$ and $n-1$ is estimated using our pre-trained optical flow network \cite{ranjan2017optical} as well as the flow between generated outputs from the same frames. The gradients from the optical flow network ($F$) are used to train the generator ($G_1$) to capture motion information and temporal continuity by minimizing the End Point Error (EPE) between the produced flows. Hence, the last component of our loss function is: \vspace{-0.05cm}%
\begin{equation}  
   \mathcal{L}_{V_n}  = ||F(G_1(x_n), G_1(x_{n-1})) - F(y_n, y_{n-1}) ||_{2},
   \label{lossTemporal}\vspace{-0.4cm}
   \end{equation}\\
where $x$ and $y$ are input and ground truth depth images respectively and $n$ the time step. While we utilize ground truth depth as inputs to the optical flow network, colour images can also be equally viable inputs. However, since our training data contains noisy environmental elements (\eg, lighting variations, rain, etc.), using the sharp and clean depth images leads to more desirable results.%
%........................................................
\begin{figure*}[t!]
	\centering
	\includegraphics[width=0.99\linewidth]{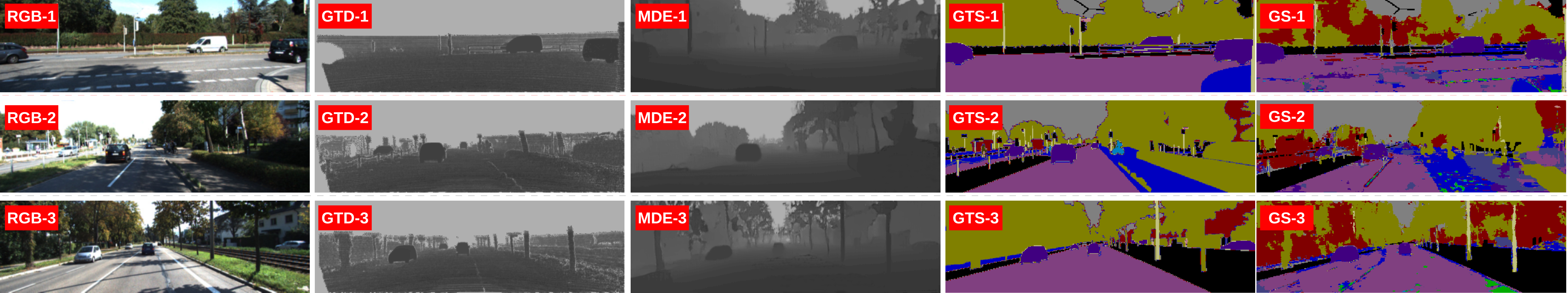}
	\captionsetup[figure]{skip=7pt}
	\captionof{figure}{Results of our approach applied to KITTI \cite{Alhaija2018IJCV, Menze2018JPRS}. \textbf{RGB:} input colour image; \textbf{GTD:} Ground Truth Depth; \textbf{MDE:} Monocular Depth Estimation; \textbf{GTS:} Ground Truth Segmentation; \textbf{GS:} Generated Segmentation.}%
	\label{fig:kitti}\vspace{-0.4cm}
\end{figure*}
%........................................................
%........................................................
\begin{figure}[t!]
	\centering
	\includegraphics[width=0.99\linewidth]{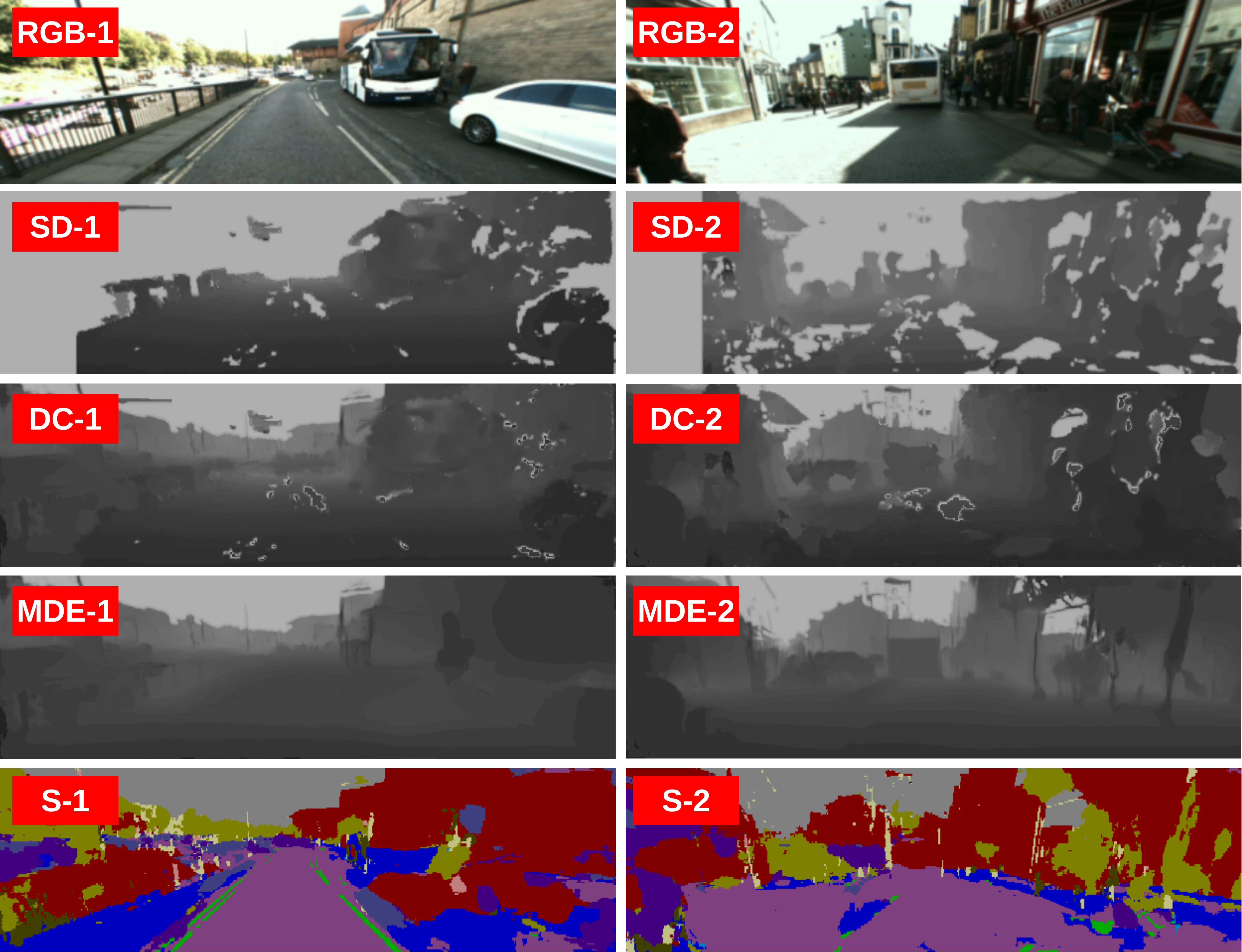}
	\captionsetup[figure]{skip=7pt}
	\captionof{figure}{Our results on locally captured data. \textbf{SD:} Depth via Stereo Correspondence; \textbf{DC:} Depth Completion; \textbf{MDE:} Monocular Depth Estimation; \textbf{S:} Semantic Segmentation.}%
	\label{fig:durham}\vspace{-0.6cm}
\end{figure}
%........................................................

Within the final decoder used exclusively for depth prediction, outputs are produced at four scales, following \cite{monodepth17}. Each scale output is twice the spatial resolution of its previous scale. The overall depth loss is therefore the sum of losses calculated at every scale $c$:\vspace{-0.3cm}%
{{\small
\begin{equation}  
\mathcal{L}_{depth} = \sum_{c=1}^{4} (\lambda_{rec}\mathcal{L}_{rec} + \lambda_{adv}\mathcal{L}_{adv} + \lambda_{s}\mathcal{L}_{s} + \lambda_{V}\mathcal{L}_{V_n}).
\label{eq:lossFinal}\vspace{-0.3cm}
\end{equation}}

The weighting coefficients ($\lambda$) are empirically selected (Section \ref{ssec:implementation_approach}). These loss components, used to optimize depth fidelity, are used alongside the semantic segmentation loss, explained in Section \ref{ssec:segmentation_approach}.%
%........................................................
\begin{table}[t!]
	\centering
	\resizebox{0.45\textwidth}{!}{
		\begin{tabular}{c c | c c}
			\hline\hline
			\multicolumn{1}{c}{Method} & \multicolumn{1}{c}{IoU} & \multicolumn{1}{c}{Method} & \multicolumn{1}{c}{IoU} \T\B \\
			\hline
			SegNet-Basic \cite{badrinarayanan2015segnet}		& 46.4 	& DeconvNet \cite{noh2015learning} 						& 48.9 				\T \\
			SegNet \cite{badrinarayanan2015segnet}				& 50.2 	& Bayesian SegNet-Basic	\cite{kendall2017bayesian}		& 55.8 				\\
			Reseg \cite{visin2016reseg}		 					& 58.8 	& Our Approach											& \textbf{59.1}		\B \\
			\hline
		\end{tabular}}
	\captionsetup[table]{skip=7pt}
	\captionof{table}{Segmentation on the CamVid \cite{brostow2009semantic} test set.}
	\label{table:camvid}
	\vspace{-0.8cm}
\end{table}
%........................................................

\subsection{Semantic Segmentation}
\label{ssec:segmentation_approach}

As semantic segmentation is not the primary focus of our approach, but only used to enforce deeper and better representation learning within our model, we opt for a simple and efficient fully-supervised training procedure for our segmentation ($G_2$). The RGB or RGB-D image is used as the input and the network outputs class labels. Pixel-wise softmax with cross-entropy is used as the loss function, with the loss summed over all the pixels within a batch:\vspace{-0.15cm}%
\begin{equation}  
P_{k}(x)  = \frac{e^{a_{k}(x)}}{\sum_{k^{\prime}=1}^{K} e^{a_{k^{\prime}}(x)}},
\label{eq:seg_1}\vspace{0.1cm}
\end{equation}
\begin{equation}  
\mathcal{L}_{seg} = - log(P_{l}(G_{2}(x))),
\label{eq:seg_2}\vspace{-0.4cm}
\end{equation}\\
where $G_{2}(x)$ denotes the network output for the segmentation task, $a_{k}(x)$ is the feature activation for channel $k$, $K$ is the number of classes, $P_{k}(x)$ is the approximated maximum function and $l$ is the ground truth label for image pixels. The loss is summed for all pixels within the images.%
%........................................................
\begin{figure*}[t!]
	\centering
	\includegraphics[width=0.99\linewidth]{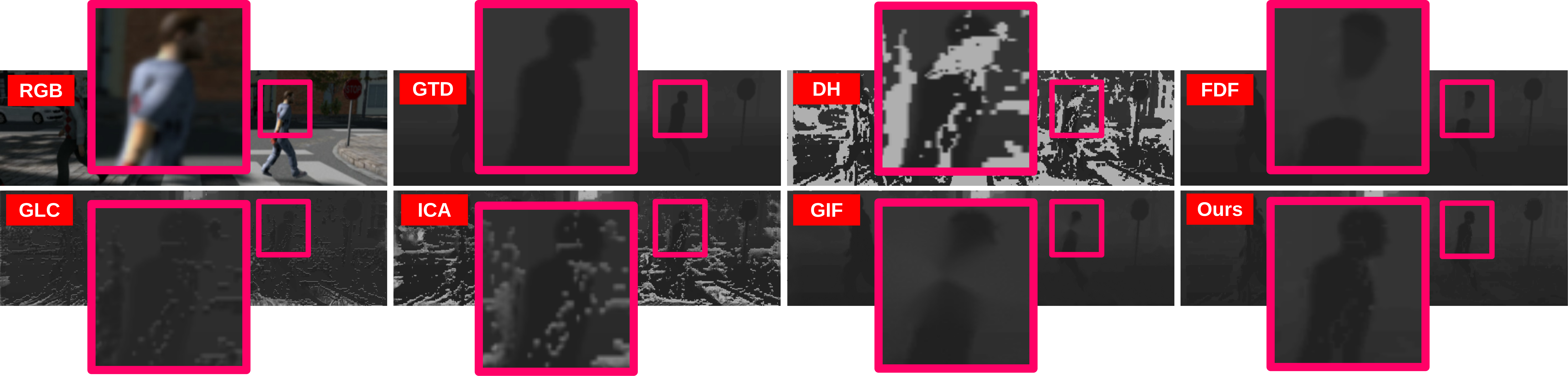}
	\captionsetup[figure]{skip=7pt}
	\captionof{figure}{Comparison of various completion methods applied to the synthetic test set. \textbf{RGB:} input colour image; \textbf{GTD:} Ground Truth Depth; \textbf{DH:} Depth Holes; \textbf{FDF:} Fourier based Depth Filling \cite{abarghouei16filling}; \textbf{GLC:} Global and Local Completion \cite{iizuka2017globally}; \textbf{ICA:} Inpainting with Contextual Attention \cite{yu2018generative}; \textbf{GIF:} Guided Inpainting and Filtering \cite{liu2012guided}.}%
	\label{fig:completion}\vspace{-0.4cm}
\end{figure*}
%........................................................

Finally, since the entire network is trained as one unit, the joint loss function is as follows:\vspace{-0.1cm}%
\begin{equation}  
	\mathcal{L}  = \mathcal{L}_{depth} + \lambda_{rec}\mathcal{L}_{seg}.
	\label{lossFinal}\vspace{-0.45cm}
\end{equation}\\
with coefficients selected empirically (Section \ref{ssec:implementation_approach}).

\subsection{Implementation Details}
\label{ssec:implementation_approach}

Synthetic data \cite{RosCVPR16} consisting of RGB, depth and class labels are used for training. The discriminator follows the architecture of \cite{radford2015unsupervised}, and the optical flow network \cite{ranjan2017optical} is pre-trained on the KITTI dataset \cite{Menze2018JPRS}. Experiments with the Sintel dataset \cite{ButlerECCV2012} returned similar, albeit slightly inferior, results. The discriminator uses convolution-BatchNorm-leaky ReLU ($slope=0.2$) modules. The dataset \cite{RosCVPR16} contains numerous sequences some spanning thousands of frames. However, a feedback network taking in high-resolution images ($512 \times 128$) back-propagating over thousands of time steps is intractable to train. Empirically, we found training over sequences of $10$ frames offers a reasonable trade-off between accuracy and training efficiency. Mini-batches are loaded in as tensors containing two sequences of $10$ frames each, resulting in roughly $10,000$ batches overall. All implementation is done in \emph{PyTorch} \cite{pytorch}, with Adam \cite{kingma2014adam} providing the best optimization ($\beta_{1} = 0.5$, $\beta_{2} = 0.999$, $\alpha = 0.0002$). The weighting coefficients in the loss function are empirically chosen to be $\lambda_{rec} = 1000, \lambda_{adv} = 100, \lambda_{s} = 10, \lambda_{V} = 1, \lambda_{seg} = 10$.\vspace{-0.1cm}

\section{Experimental Results}
\label{sec:results}\vspace{-0.1cm}

We assess our approach using ablation studies and both qualitative and quantitative comparisons with state-of-the-art methods applied to publicly available datasets \cite{Alhaija2018IJCV, brostow2009semantic, cordts2016cityscapes, Geiger2013IJRR, Menze2018JPRS}. We also utilize our own synthetic test set and data captured locally to further evaluate the approach.%

\subsection{Ablation Studies}
\label{ssec:ablation}\vspace{-0.1cm}

A crucial part of our work is demonstrating that every component of the approach is integral to the overall performance. We train our model to perform two tasks based on the assumption that the network is forced to learn more about the scene if different objectives are to be accomplished. We demonstrate this by training one model performing both tasks and two separate models focusing on each and conducting tests on randomly selected synthetic sequences \cite{RosCVPR16}. As seen in Table \ref{table:ablation_separation}, both tasks (monocular depth estimation and semantic segmentation) perform better when the model is trained on both. Moreover, since the segmentation pipeline does not receive any explicit temporal supervision (from the optical flow network) and its temporal continuity is only enforced by the input and middle streams trained by the depth pipeline, when the two pipelines are disentangled, the segmentation results become far worse than the depth results (Table \ref{table:ablation_separation}).%
%........................................................
\begin{table}[t!]
	\centering
	\resizebox{0.48\textwidth}{!}{
		\begin{tabular}{l c c | l c c}
			\hline\hline
			\multicolumn{1}{c}{Method} & \multicolumn{1}{c}{PSNR} & \multicolumn{1}{c}{SSIM} &\multicolumn{1}{c}{Method} & \multicolumn{1}{c}{PSNR} & \multicolumn{1}{c}{SSIM} \T\B \\
			\hline
			Holes								& 33.73			 	& 0.372	&	GTS \cite{iizuka2017globally}		& 31.47 			& 0.672				\T \\
			ICA \cite{yu2018generative}			& 31.01 			& 0.488	&	GIF \cite{liu2012guided}			& 44.57 			& 0.972				\\			
			FDF \cite{abarghouei16filling}		& 46.13			 	& 0.986 &	Ours 								& \textbf{47.45} 	& \textbf{0.991} 	\B \\
			\hline
		\end{tabular}}
	\captionsetup[table]{skip=7pt}
	\captionof{table}{Structural integrity analysis post depth completion.}
	\label{table:completion}
	\vspace{-0.6cm}
\end{table}
%........................................................

Figure \ref{fig:temporal} depicts the quality of the outputs when the model is a feedback network trained temporally compared to our model when the output depth from the previous time step is not used as the input during training. We can clearly see that both depth and segmentation results are of higher fidelity when temporal information is used during training.%
%............................................................................
\begin{table*}[!t]
	\centering
	\resizebox{\textwidth}{!}{
		{\tabulinesep=0mm
			\begin{tabu}{@{\extracolsep{5pt}}l c c c c c c c c@{}}
				\hline\hline
				\multicolumn{1}{l}{\multirow{2}{*}{Method}} & 
				\multicolumn{1}{c}{\multirow{2}{*}{}} & 
				\multicolumn{4}{c}{Error Metrics (lower, better)} & 
				\multicolumn{3}{c}{Accuracy Metrics (higher, better)} \T\B \\
				\cline{3-6} \cline{7-9}
				&						& Abs. Rel.		& Sq. Rel. 		& RMSE 			& RMSE log 		& $\sigma < 1.25$& $\sigma < 1.25^{2}$& $\sigma < 1.25^{3}$ \T\B \\
				\hline\hline
				Train Set Mean 		& \cite{Geiger2013IJRR}		 	& 0.403 		& 0.530 		& 8.709 		& 0.403 		& 0.593			& 0.776 		& 0.878 \T \\
				Eigen \etal		 	& \cite{eigen2014depth} 		& 0.203 		& 1.548 		& 6.307 		& 0.282 		& 0.702 		& 0.890 		& 0.958 \\
				Liu \etal 			& \cite{liu2016learning}		& 0.202			& 1.614 		& 6.523 		& 0.275 		& 0.678 		& 0.895 		& 0.965 \\
				Zhou \etal 			& \cite{zhou2017unsupervised}	& 0.208 		& 1.768 		& 6.856 		& 0.283 		& 0.678 		& 0.885 		& 0.957 \\
				Godard \etal 		& \cite{monodepth17} 			& 0.148 & \textbf{1.344}		& 5.927 		& 0.247 		& 0.803 		& 0.922 		& 0.964 \\
				Zhan \etal			& \cite{zhan2018unsupervised}	& \textbf{0.144}&1.391			& \textbf{5.869}& 0.241			& 0.803			& 0.928			& \textbf{0.969} \B \\
				\hline
				Our Approach 		& 		& 0.193 		& 1.438									& 5.887& \textbf{0.234}& \textbf{0.836}& \textbf{0.930}& 0.958 \T\B \\
				\hline
			\end{tabu}}}
	\captionsetup[table]{skip=7pt}
	\captionof{table}{Numerical comparison of monocular depth estimation over the KITTI \cite{Geiger2013IJRR} data split in \cite{eigen2014depth}. All comparators are trained and tested on the same dataset (KITTI \cite{Geiger2013IJRR}) while our approach is trained on \cite{RosCVPR16} and tested using \cite{Geiger2013IJRR}.}
	\label{table:compare_MDE}\vspace{-0.5cm}
\end{table*}
%............................................................................

Additionally, our depth prediction pipeline uses several loss functions. We employ the same test sequences to evaluate our model trained as different components are removed. Table \ref{table:ablation_remove_components} demonstrates the network temporally trained with all the loss components (T/R/A/SC/S/OF) outperforms models trained without specific ones. Qualitatively, we can see in Figure \ref{fig:ablation} that the results are far better when the network is fully trained with all the components. Specifically, the set of skip connections used in the network make a significant difference in the quality of the outputs.\vspace{-0.1cm}%

\subsection{Semantic Segmentation}
\label{ssec:segmentation_results}\vspace{-0.1cm}

Segmentation is not the focus of this work and is mainly used to boost the performance of depth prediction. However, we extensively evaluate our segmentation pipeline which outperforms several well-known comparators. We utilize Cityscapes \cite{cordts2016cityscapes} and CamVid \cite{brostow2009semantic} test sets for our performance evaluation despite the fact that our model is solely trained on synthetic data and \textit{without any domain adaptation} should not be expected to perform well on naturally sensed real-world data. The effective performance of our segmentation points to the generalization capabilities of our model. When tested on CamVid \cite{brostow2009semantic}, our approach produces better results compared to well-established techniques such as \cite{badrinarayanan2015segnet, kendall2017bayesian, noh2015learning, visin2016reseg} despite the lower quality of the input images as seen in Table \ref{table:camvid}. As for Cityscapes \cite{cordts2016cityscapes}, the test set does not contain video sequences, but our temporal model still outperforms approaches such as \cite{chen2018deeplab, liu2015semantic, long2015fully, uhrig2016pixel, zheng2015conditional}, as demonstrated in Table \ref{table:cityscapes}.%

Examples of the segmentation results over both datasets are seen in Figure \ref{fig:segmentation}. Additionally, we also use the KITTI semantic segmentation data \cite{Alhaija2018IJCV} in our tests and as shown in Figure \ref{fig:kitti}, our approach produces high fidelity semantic class labels despite including \textit{no domain adaptation}.\vspace{-0.2cm}%

\subsection{Depth Completion}
\label{ssec:completion_results}\vspace{-0.1cm}

Evaluation for depth completion ideally requires dense ground truth scene depth. However, no such dataset exists for urban driving scenarios, which is why we utilize randomly selected previously unseen synthetic data with available dense depth images to assess the results. Our model generates full scene depth and the predicted depth values for the missing regions of the depth image are subsequently blended in with the known regions of the image using \cite{perez2003poisson}. Figure \ref{fig:completion} shows a comparison of our results against other contemporary approaches \cite{abarghouei16filling, iizuka2017globally, liu2012guided, yu2018generative}. As seen from the enlarged sections, our approach produces minimal artefacts (blurring, streaking, etc.) compared to the other techniques. To evaluate the structural integrity of the results post completion, we also numerically assess the performance of our approach and the comparators. As seen in Table \ref{table:completion}, our approach quantitatively outperforms the comparators as well.%
%........................................................
\begin{figure*}[t!]
	\centering
	\includegraphics[width=0.99\linewidth]{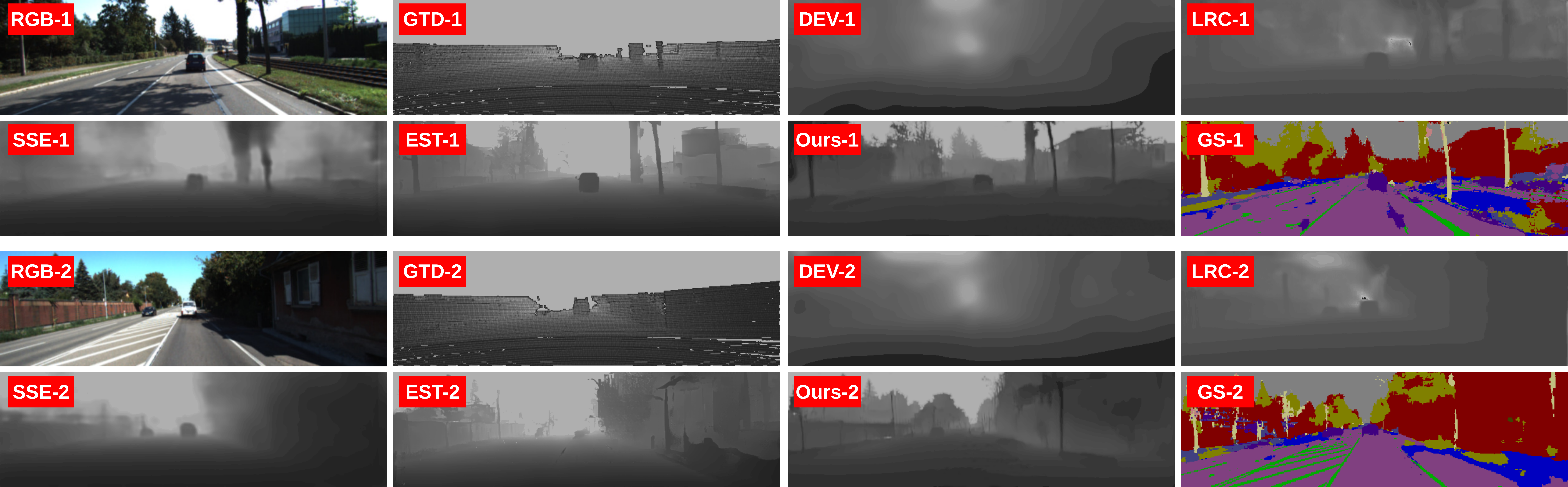}
	\captionsetup[figure]{skip=7pt}
	\captionof{figure}{Comparing the results of the approach against \cite{zhou2017unsupervised, monodepth17, kuznietsov2017semi, atapour2018real}. Images have been adjusted for better visualization. \textbf{RGB:} input colour image;  \textbf{GTD:}  Ground Truth Depth; \textbf{DEV:} Depth and Ego-motion from Video \cite{zhou2017unsupervised}; \textbf{LRC:} Left-Right Consistency \cite{monodepth17}; \textbf{SSE:} Semi-supervised Estimation \cite{kuznietsov2017semi}; \textbf{EST:} Estimation via Style Transfer \cite{atapour2018real}; \textbf{GS:} Generated Segmentation.}%
	\label{fig:estimation}\vspace{-0.6cm}
\end{figure*}
%........................................................

While blending \cite{perez2003poisson} might work well for colour images with a connected missing region, significant quantities of small and large holes in depth images can lead to undesirable artefacts such as stitch mark or burning effects post blending. Examples of artefacts can be seen in Figure \ref{fig:durham}, which demonstrates the results of the approach applied to locally captured data. This is further discussed in Section \ref{sec:limits}\vspace{0.0cm}.%

\subsection{Monocular Depth Estimation}
\label{ssec:estimation_results}\vspace{0.0cm}

As the main focus of our model, our monocular depth estimation model is evaluated against contemporary state-of-the-art approaches \cite{atapour2018real, eigen2014depth, monodepth17, liu2016learning, zhan2018unsupervised, zhou2017unsupervised}. Following the conventions of the literature, we use the data split suggested in \cite{eigen2014depth} as the test set. These images are selected from random sequences and do not follow a temporally sequential pattern, while our full approach requires video sequences as its input. As a result, we apply our approach to all the sequences from which the images are chosen but the evaluation itself is only performed on the 697 test images.%

For numerical assessment, the generated depth is corrected for the differences in focal length between the training \cite{RosCVPR16} and testing data \cite{Geiger2013IJRR}. As seen in Table \ref{table:compare_MDE}, our approach outperforms \cite{eigen2014depth, liu2016learning, zhou2017unsupervised} across all metrics and stays competitive with \cite{monodepth17, zhan2018unsupervised}. It is important to note that all of these comparators are trained on the \emph{same} dataset as the one used for testing \cite{Geiger2013IJRR} while our approach is trained on synthetic data \cite{RosCVPR16} \emph{without domain adaptation} and has not seen a single image from \cite{Geiger2013IJRR}. Additionally, none of the other comparators is capable of producing temporally consistent outputs as all of them operate on a frame level. As this cannot be readily illustrated via still images within Figures \ref{fig:completion} and \ref{fig:estimation}, we kindly invite the reader to view the supplementary \href{https://vimeo.com/325161805}{\textbf{video}} material accompanying the paper.%

We also assess our model using the data split of KITTI \cite{Menze2018JPRS} and qualitatively evaluate the results, since the ground truth images in \cite{Menze2018JPRS} are of higher quality than the laser data and provide CAD models as replacements for the cars in the scene. As shown in Figure \ref{fig:kitti}, our method produces sharp and crisp depth outputs with segmentation results in which object boundaries and thin structures are well preserved.\vspace{-0.1cm}

\section{Limitations and Future Work}
\label{sec:limits}\vspace{-0.05cm}

Even though our approach can generate temporally consistent depth and segmentation by utilizing a feedback network, this can lead to error propagation, \ie, when an erroneous output is generated at one time step, the invalid values will continually propagate to future frames. This can be resolved by exploring the use of 3D convolutions or regularization terms aimed at penalizing propagated invalid outputs. Moreover, as mentioned in Section \ref{ssec:completion_results}, blending the depth output into the known regions of the depth \cite{perez2003poisson} produces undesirable artefacts in the results. This can be rectified by incorporating the blending operation into the training procedure. In other words, the blending itself will take place before the supervisory signal is back-propagated through the network during training, which would force the network to learn these artefacts, removing any need for post-processing. As for our segmentation component, no explicit temporal consistency enforcement or class balancing is performed, which has lead to frame-to-frame flickering and lower accuracy with unbalanced classes (\eg, pedestrians, cyclists). By improving segmentation, the entire model can benefit from a performance boost. Most of all, the use of domain adaptation \cite{atapour2018real, hoffman2018cycada} can significantly improve all results since despite its generalization capabilities, the model is only trained on synthetic data and should not be expected to perform just as well on naturally-sensed real-world images.\vspace{-0.1cm}% 

\section{Conclusion}
\label{sec:conclusion}\vspace{-0.1cm}

We propose a multi-task model capable of performing depth prediction and semantic segmentation in a temporally consistent manner using a feedback network that takes as its recurrent input the output generated at the previous time step. Using a series of dense skip connections, we ensure that no high-frequency spatial information is lost during feature down-sampling within the training process. We consider the task of depth prediction within the areas of depth completion and monocular depth estimation, and therefore train models based on both objectives within the depth prediction component. Using extensive experimentation, we demonstrate that our model achieves much better results when it performs depth prediction and segmentation at the same time compared to two separate networks performing the same tasks. The use of skip connections is also shown to be significantly effective in improving the results for both depth prediction and segmentation tasks. Although certain isolated issues remain, experimental evaluation demonstrates the efficacy of our approach compared to contemporary state-of-the-art methods tackling the same problem domains \cite{chen2018deeplab, monodepth17, iizuka2017globally, kendall2017bayesian, long2015fully, yu2018generative, zhan2018unsupervised, zhou2017unsupervised}.%
\\
\textit{We also invite the readers to refer to the \href{https://vimeo.com/325161805}{\textbf{video}}: \href{https://vimeo.com/325161805}{https://vimeo.com/325161805} and the \href{https://github.com/atapour/temporal-depth-segmentation}{source code}.}

{\small
\bibliographystyle{ieee_fullname}
\bibliography{main-bib}
}

\section{Appendix}

In this section, we provide additional information that could not be placed within the main paper due to space restrictions. We kindly invite the readers to watch the \href{https://vimeo.com/325161805}{\textbf{video}} submitted as part of the supplementary material along with this document.

%-------------------------------------------------------------------------
\subsection{Hole Prediction Network}

As mentioned in the main manuscript, when the model is being trained to perform depth completion, the input must be a four-channel RGB-D image, in which the depth channel contains holes that would naturally occur when sensed through imperfect capture technologies. However, the dataset used for training our model \cite{RosCVPR16} consists of pixel-prefect depth images without any holes.% 

This synthetic dataset \cite{RosCVPR16} does contain stereo image pairs, so a simple solution would be to calculate the disparity and subsequently the depth using a well-established stereo matching approach such as Semi-Global Matching \cite{hirschmuller2008stereo} and use the resulting depth image (which will contain holes) as the input.

However, each image in a stereo pair in \cite{RosCVPR16} (left and right) comes with its own corresponding (left and right) depth image, and half of the dataset (aligned RGB and depth images) will be rendered useless if stereo matching is used to calculate depth images with hole.%

As a result, we opt for training an entirely separate model that would be responsible for creating holes in the depth images. Even though the details regarding the training or use of this network have no bearing on the approach proposed in the main manuscript, we will attempt to cover the inner workings and experimental evaluation of our \textit{hole prediction} model here.

This \textit{hole prediction} model is a fully convolutional encoder-decoder network inspired by \cite{ronneberger2015u} with skip connections between all corresponding layers in the encoder and the decoder. The last decoder layer is connected to a soft-max classifier. Each convolutional layer is followed by batch normalization \cite{ioffe2015batch} and a ReLU. The network architecture can be seen in Figure \ref{fig:hole_pred_arch_app}.

%........................................................
\begin{figure*}[b!]
	\centering
	\includegraphics[width=0.99\linewidth]{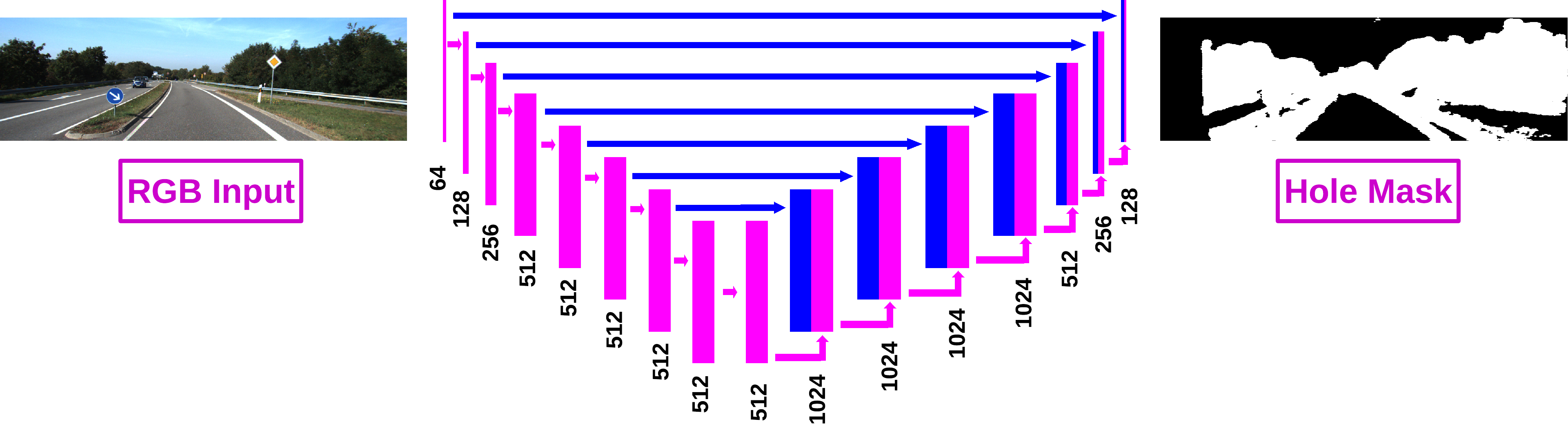}
	\captionsetup[figure]{skip=7pt}
	\captionof{figure}{Overview of the architecture of the \emph{hole prediction} network.}%
	\label{fig:hole_pred_arch_app}\vspace{-0.4cm}
\end{figure*}
%........................................................

The training data for this \textit{hole prediction} network is made up of $30,000$ pairs of stereo images from \cite{Geiger2013IJRR}. Disparity is calculated using Semi-Global Matching (SGM) \cite{hirschmuller2008stereo} and a hole mask ($M$) is subsequently generated which indicates which pixels are holes. Although SGM is used here, this is interchangeable with any other passive or active depth capture approach. The left RGB images are thus used as inputs with the generated masks as ground truth labels. Binary cross-entropy is used as the loss function since the segmentation task involves only two classes: hole and non-hole.

Qualitative analyses reveal that holes are predicted where expected. From Figure \ref{fig:hole_pred_imgs_app}, we see that in regions where camera overlap is absent or featureless surfaces, sparse shrubbery, unclear object boundaries, and very distant objects are present, such pixels are correctly classified as holes.
%........................................................
\begin{figure*}[b!]
	\centering
	\includegraphics[width=0.99\linewidth]{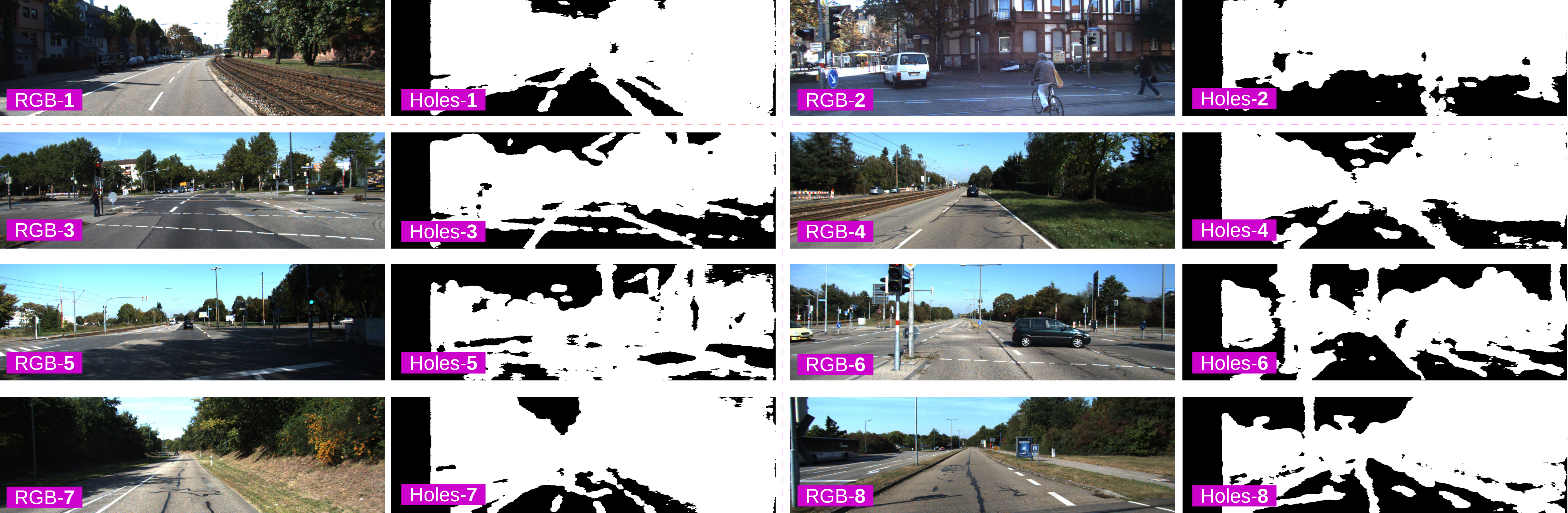}
	\captionsetup[figure]{skip=7pt}
	\captionof{figure}{Examples of results of the \textit{hole prediction} model applied to unseen images from \cite{Geiger2013IJRR}.}%
	\label{fig:hole_pred_imgs_app}\vspace{-0.4cm}
\end{figure*}
%........................................................

\subsection{Additional Experiments}

Following the conventions of the expansive literature on monocular depth estimation, we measure the performance of our approach against the KITTI dataset \cite{Geiger2013IJRR}. However, we have re-trained and tested all the comparators using the synthetic dataset of \cite{RosCVPR16} but for brevity and due to our superior performance on the \emph{unseen} KITTI dataset, against comparators \emph{actually} trained on KITTI, we have not included these extra results in the main manuscript. Table \ref{table:comapre_synth} presents the comparison of our approach against \cite{monodepth17, zhou2017unsupervised} trained on the synthetic dataset of \cite{RosCVPR16} under the exact same conditions as outlined in Section 3.4 of the main manuscript. Our approach outperforms the comparators by a large margin (Table \ref{table:comapre_synth}).
%............................................................................
\begin{table}[!t]
	\centering
	\resizebox{\columnwidth}{!}{
		{\tabulinesep=0mm
			\begin{tabu}{@{\extracolsep{5pt}}c c c c c c}
				\hline\hline
				\multicolumn{1}{l}{\multirow{2}{*}{Method}} &  
				\multicolumn{4}{c}{Error} & 
				\multicolumn{1}{c}{Accuracy} \T\B \\
				\cline{2-5} \cline{6-6}
							   & Abs. Rel.		& Sq. Rel.		   & RMSE			& RMSE log		&  $\sigma < 1.25^{3}$ \T\B \\
				\hline\hline
				\cite{zhou2017unsupervised}	& 0.401 		   & 1.601 		      & 6.598 		   & 0.363 		   & 	0.788		 		\T \\
            	\cite{monodepth17}	& 0.334 		   & 1.556 		      & 6.304 		   & 0.302 		   &  0.852 					\\
            \hline
				Ours (full) & \textbf{0.208}& \textbf{1.402}& \textbf{6.026}& \textbf{0.269}& \textbf{0.926}  \T\B \\
				\hline
			\end{tabu}}}
	\captionsetup[table]{skip=7pt}
	\captionof{table}{\small{Comparisons using synthetic data \cite{RosCVPR16}.}}
	\label{table:comapre_synth}\vspace{-0.65cm}
\end{table}
%............................................................................

\subsection{Figures}

Due to the space restrictions, some of the figures within the main paper may be too small for appropriate viewing. While some of the results are better seen in the accompanying \href{https://vimeo.com/325161805}{video}, we also provide enlarged versions of some of the figures here. Please see Figures \ref{fig:pipline_app}, \ref{fig:architecture_app}, \ref{fig:ablation_app}, \ref{fig:temporal_app}, \ref{fig:segmentation_app} and \ref{fig:completion_app} in the appendix.
\\
Video URL: \href{https://vimeo.com/325161805}{https://vimeo.com/325161805}
%........................................................
\begin{figure*}[t!]
	\centering
	\includegraphics[width=0.99\linewidth]{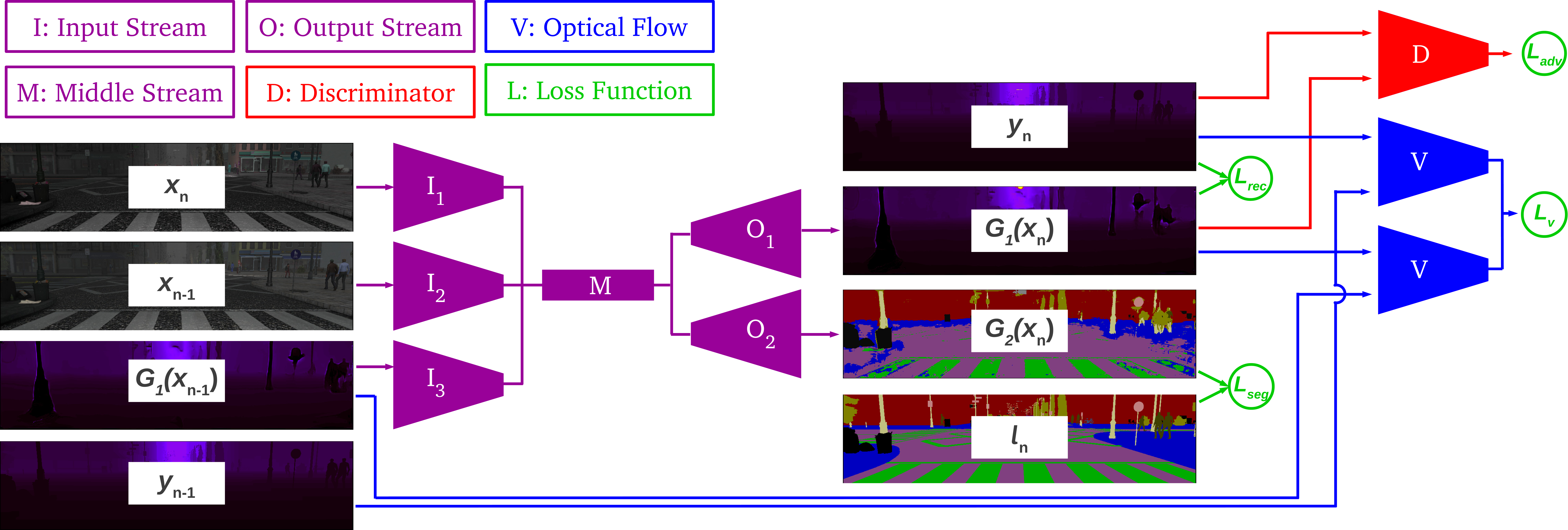}
	\captionsetup[figure]{skip=7pt}
	\captionof{figure}{An outline of the training procedure of the main proposed approach.}%
	\label{fig:pipline_app}\vspace{-0.4cm}
\end{figure*}
%........................................................
%........................................................
\begin{figure*}[t!]
	\centering
	\includegraphics[width=0.99\linewidth]{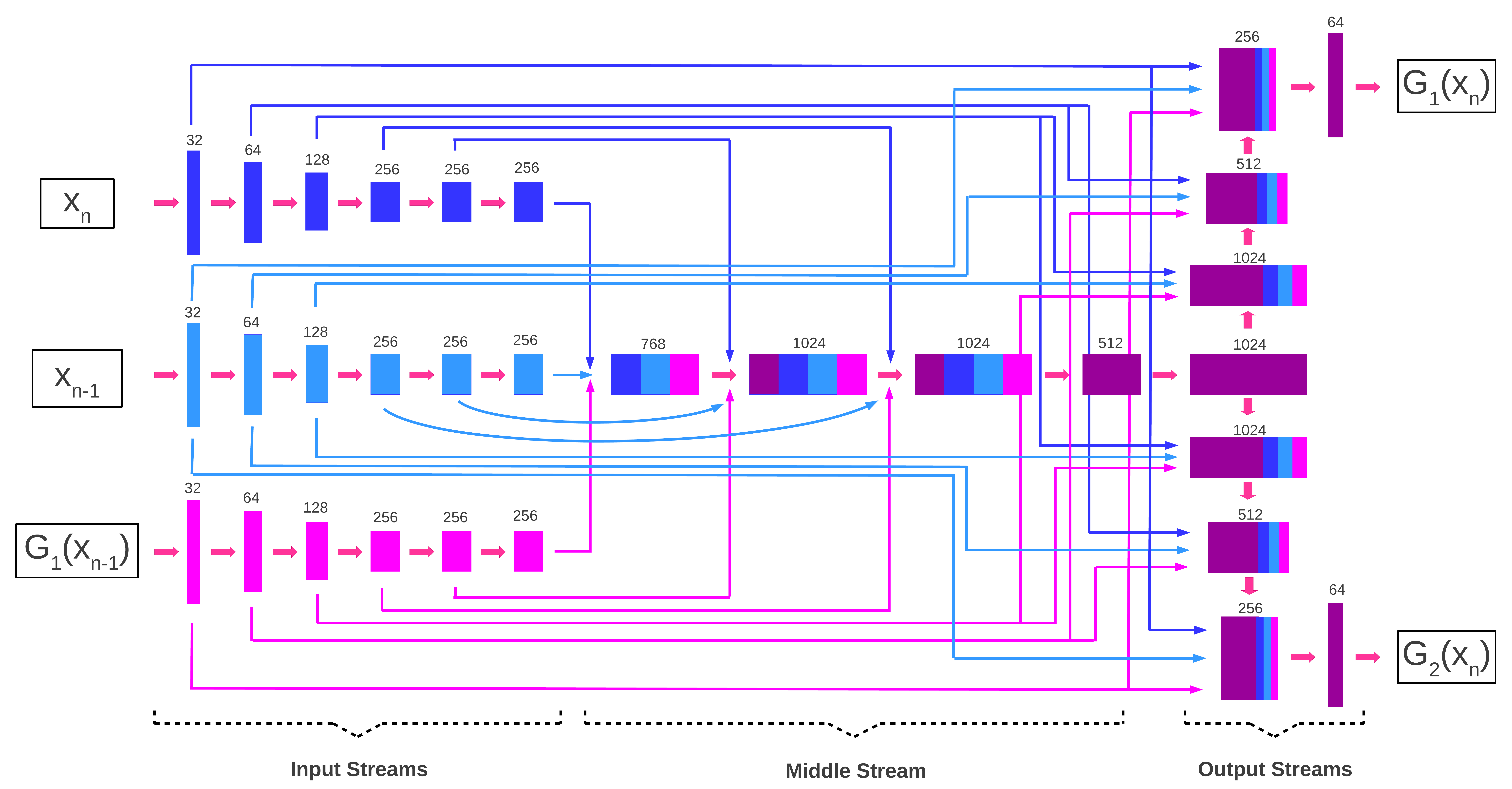}
	\captionsetup[figure]{skip=7pt}
	\captionof{figure}{An overview of the generator architecture.}%
	\label{fig:architecture_app}\vspace{-0.4cm}
\end{figure*}
%........................................................
%........................................................
\begin{figure*}[t!]
	\centering
	\includegraphics[width=0.99\linewidth]{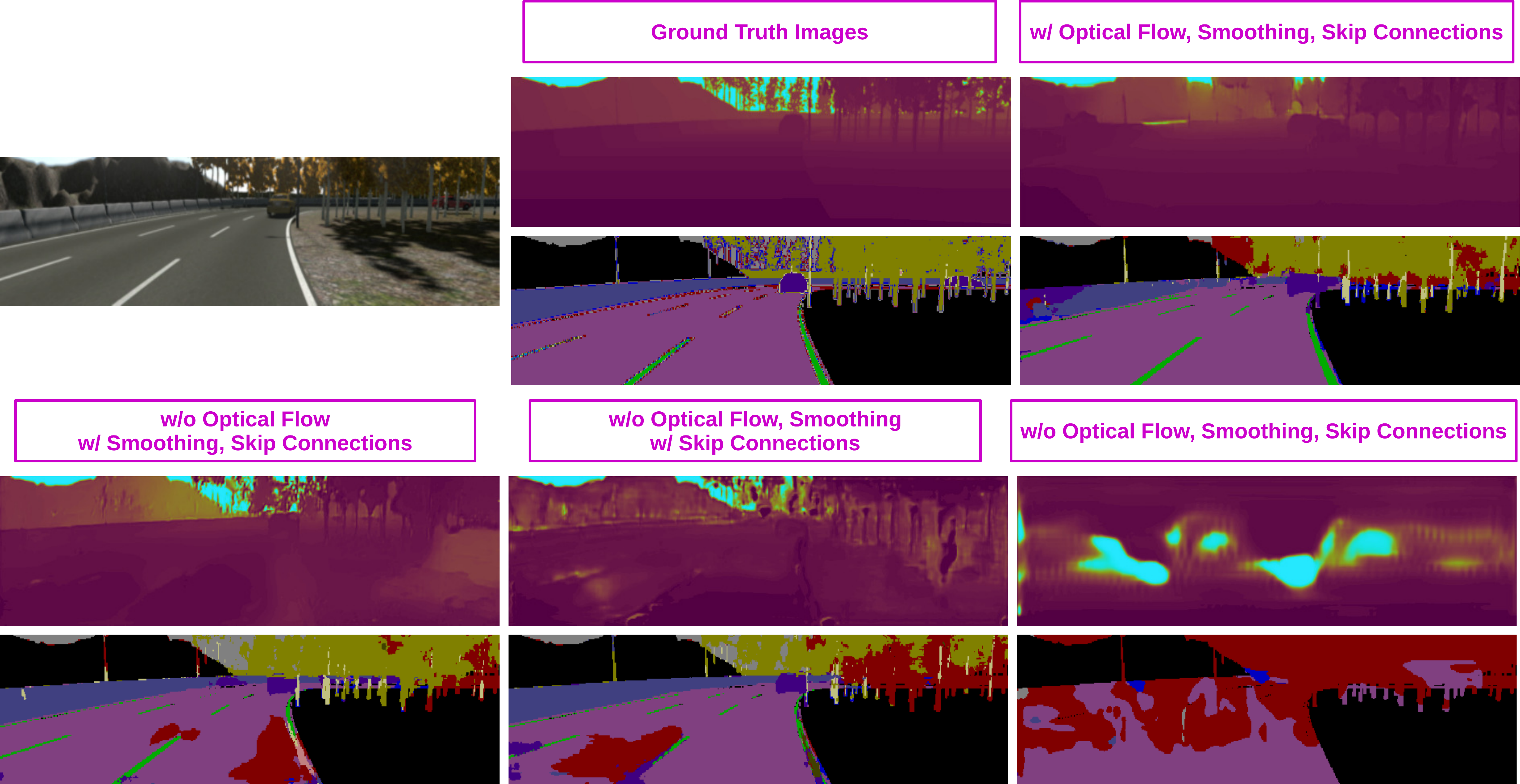}
	\captionsetup[figure]{skip=7pt}
	\captionof{figure}{Comparing the results of our model (with monocular depth estimation) when different components of the approach are removed.}%
	\label{fig:ablation_app}\vspace{-0.4cm}
\end{figure*}
%........................................................
%........................................................
\begin{figure*}[t!]
	\centering
	\includegraphics[width=0.99\linewidth]{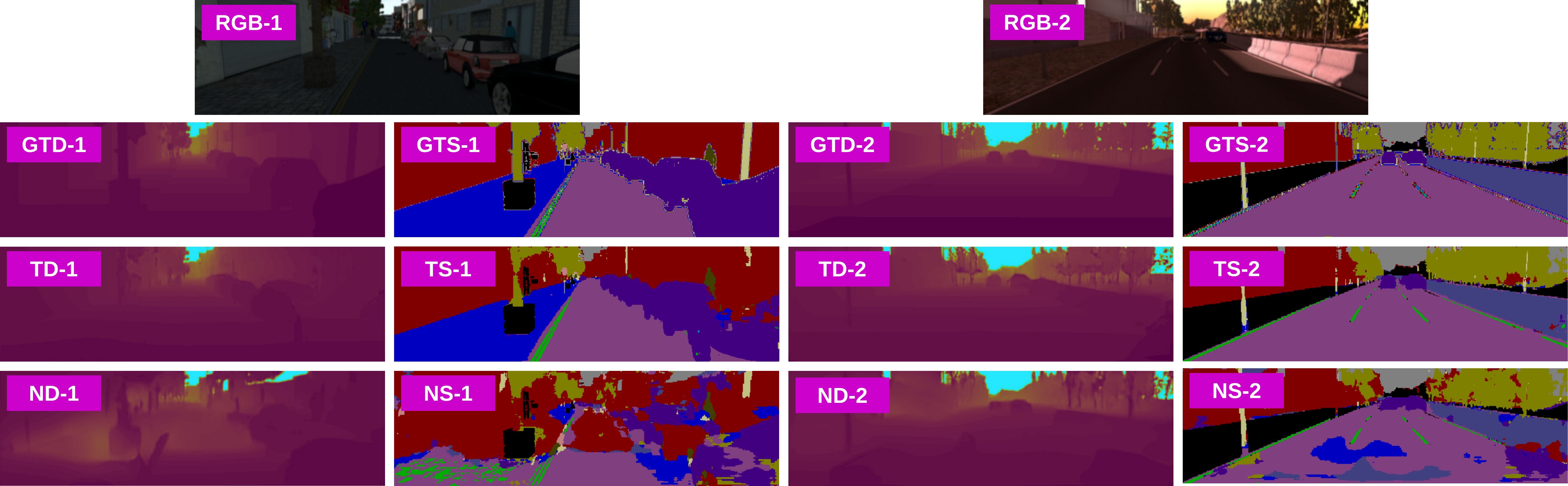}
	\captionsetup[figure]{skip=7pt}
	\captionof{figure}{Comparing the results of the approach on the synthetic test set when the model is trained with and without temporal consistency. \textbf{RGB:} input colour image; \textbf{GTD:} Ground Truth Depth; \textbf{GTS:} Ground Truth Segmentation; \textbf{TS:} Temporal Segmentation; \textbf{TD:} Temporal Depth; \textbf{NS:} Non-Temporal Segmentation; \textbf{ND:} Non-Temporal Depth.}%
	\label{fig:temporal_app}\vspace{-0.4cm}
\end{figure*}
%........................................................
%........................................................
\begin{figure*}[t!]
	\centering
	\includegraphics[width=0.99\linewidth]{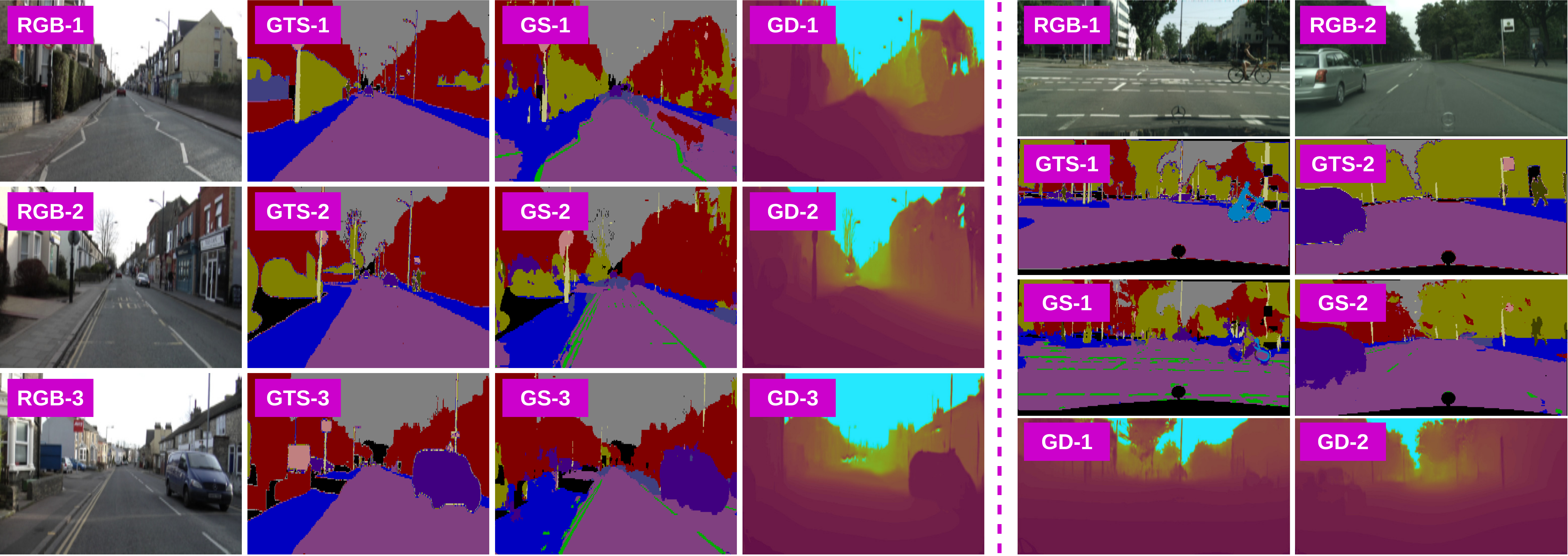}
	\captionsetup[figure]{skip=7pt}
	\captionof{figure}{Results of our approach on CamVid \cite{brostow2009semantic} (left) and Cityscapes \cite{cordts2016cityscapes} (right) datasets. \textbf{RGB:} input colour image; \textbf{GTS:} Ground Truth Segmentation; \textbf{GS:} Generated Segmentation; \textbf{GD:} Generated Depth.}%
	\label{fig:segmentation_app}\vspace{-0.3cm}
\end{figure*}
%........................................................
%........................................................
\begin{figure*}[t!]
	\centering
	\includegraphics[width=0.99\linewidth]{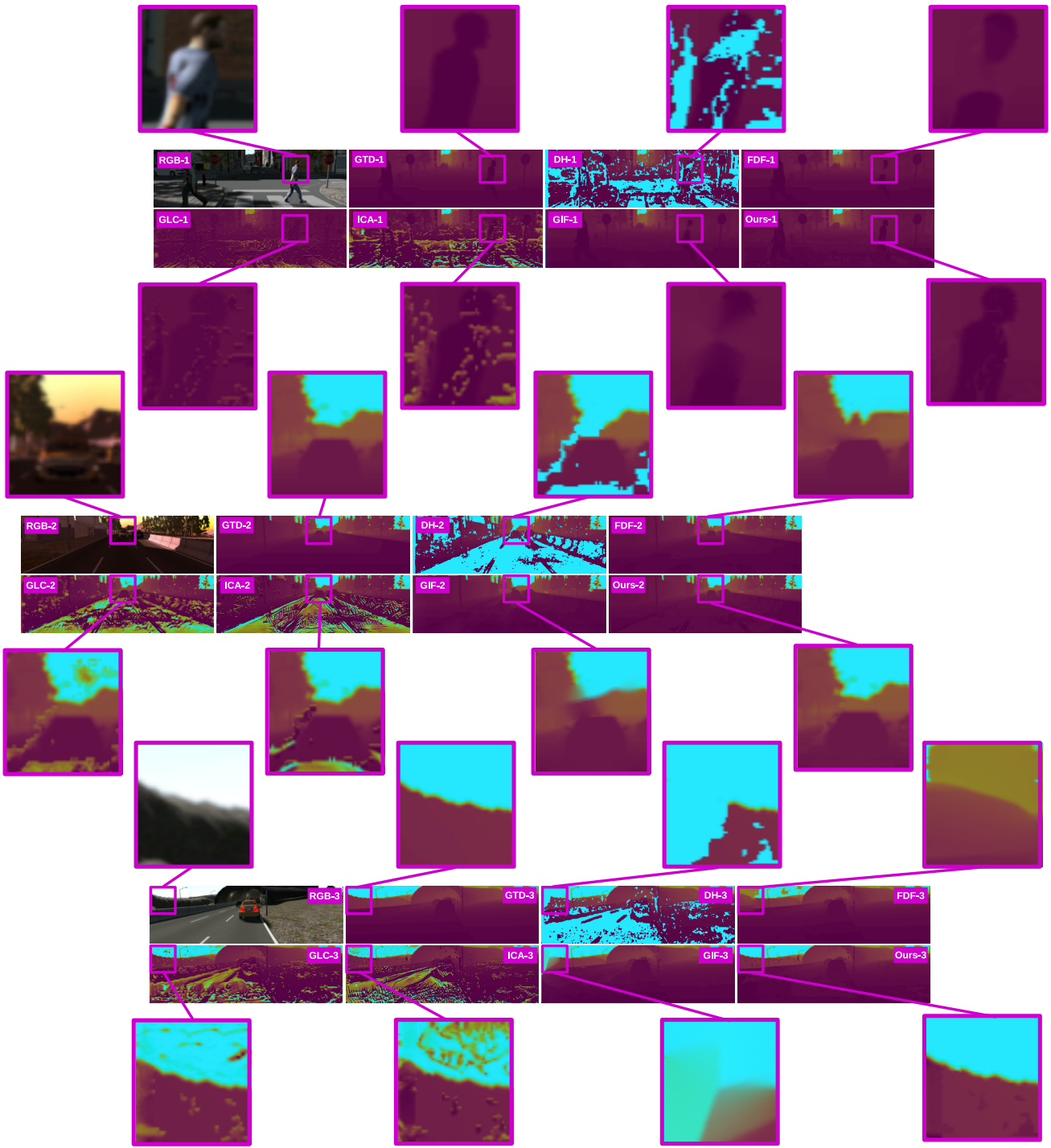}
	\captionsetup[figure]{skip=7pt}
	\captionof{figure}{Comparison of depth completion methods applied to synthetic test set. \textbf{RGB:} input colour image; \textbf{GTD:} Ground Truth Depth; \textbf{DH:} Depth Holes; \textbf{FDF:} Fourier based Depth Filling \cite{abarghouei16filling}; \textbf{GTS:} Global and Local Completion \cite{iizuka2017globally}; \textbf{ICA:} Inpainting with Contextual Attention \cite{yu2018generative}; \textbf{GIF:} Guided Inpainting and Filtering \cite{liu2012guided}.}%
	\label{fig:completion_app}\vspace{-0.4cm}
\end{figure*}
%........................................................

\end{document}